\documentclass[runningheads,a4paper]{llncs}

\usepackage{amssymb}
\usepackage{graphicx}
\usepackage{subfig}

\usepackage{url}
\urldef{\mailsa}\path|{alfred.hofmann, ursula.barth, ingrid.haas, frank.holzwarth,|
\urldef{\mailsb}\path|anna.kramer, leonie.kunz, christine.reiss, nicole.sator,|
\urldef{\mailsc}\path|erika.siebert-cole, peter.strasser, lncs}@springer.com|    
\newcommand{\keywords}[1]{\par\addvspace\baselineskip
\noindent\keywordname\enspace\ignorespaces#1}

\begin{document}

\mainmatter  

\title{3D FCN Feature Driven Regression Forest-Based Pancreas Localization and Segmentation}

\titlerunning{3D FCN Feature Driven Pancreas Localization and Segmentation}

%
%
\author{Masahiro Oda \inst{1}
\and
Natsuki Shimizu \inst{2}
\and
Holger R. Roth \inst{1}
\and
Ken'ichi Karasawa \inst{2}
\and
Takayuki Kitasaka \inst{3}
\and
Kazunari Misawa \inst{4}
\and
Michitaka Fujiwara \inst{5}
\and \\
Daniel Rueckert \inst{6}
\and
Kensaku Mori \inst {7,1}}

\authorrunning{Masahiro Oda et al.}


\institute{Graduate School of Informatics, Nagoya University,\\
\email{moda@mori.m.is.nagoya-u.ac.jp}\\
\and
Graduate School of Information Science, Nagoya University,\\
\and
School of Information Science, Aichi Institute of Technology,\\
\and
Aichi Cancer Center,\\
\and
Nagoya University Graduate School of Medicine,\\
\and
Department of Computing, Imperial College London,\\
\and
Strategy Office, Information and Communications, Nagoya University,\\
}

%
%

\toctitle{Lecture Notes in Computer Science}
\tocauthor{Authors' Instructions}
\maketitle

\begin{abstract}
This paper presents a fully automated atlas-based pancreas segmentation method from CT volumes utilizing 3D fully convolutional network (FCN) feature-based pancreas localization.
Segmentation of the pancreas is difficult because it has larger inter-patient spatial variations than other organs.
Previous pancreas segmentation methods failed to deal with such variations.
We propose a fully automated pancreas segmentation method that contains novel localization and segmentation.
Since the pancreas neighbors many other organs, its position and size are strongly related to the positions of the surrounding organs.
We estimate the position and the size of the pancreas (localized) from global features by regression forests.
As global features, we use intensity differences and 3D FCN deep learned features, which include automatically extracted essential features for segmentation.
We chose 3D FCN features from a trained 3D U-Net, which is trained to perform multi-organ segmentation.
The global features include both the pancreas and surrounding organ information.
After localization, a patient-specific probabilistic atlas-based pancreas segmentation is performed.
In evaluation results with 146 CT volumes, we achieved 60.6\% of the Jaccard index and 73.9\% of the Dice overlap.
\keywords{Segmentation, Pancreas, Fully convolutional network, Regression forest}
\end{abstract}

\section{Introduction}

Medical images contain much useful information for computer-aided diagnosis and interventions.
Organ segmentation from medical images is necessary to utilize such information for many applications.
Many abdominal organ segmentation methods have been reported \cite{Okada08,Chu13,Wolz13,Karasawa15,Tong15,Saito16,Roth16,Oda16}.
We used the pancreas segmentation results obtained from them in diagnosis, visualization, surgical planning, and surgical simulation.
Pancreas segmentation accuracies from CT volumes are commonly lower than those of other abdominal organs in previous methods.
The main cause of the difficulty comes from the large spatial (position and size) variation of the pancreas.
Because it is small and neighbors many organs, its actions and shape are influenced by the movements and the deformations of its neighbor organs.
Therefore, prior to pancreas segmentation, pancreas localization using global information, which might include the intensity, texture, appearance, or structure of the surrounding organs, should be performed for higher segmentation accuracies.
Pancreas localization denotes a rough estimation of the pancreas position and size.

Most previous pancreas segmentation methods employ probabilistic atlas-based segmentation approaches \cite{Okada08,Chu13,Wolz13,Karasawa15,Tong15,Oda16}.
A patient-specific-atlas \cite{Okada08,Karasawa15,Tong15,Oda16} and a hierarchical-atlas \cite{Wolz13} have been used.
However, such atlas-based approaches cannot deal with an organ's spatial variation.
Some methods localized pancreas atlases using manually specified information \cite{Okada08,Karasawa15} or the segmentation results of other organs \cite{Chu13,Wolz13}.
These methods depend on other organ segmentation results or require manual interactions.
Recently, pancreas segmentation methods have been proposed with automated localization techniques \cite{Roth16,Oda16}.
Roth et al. \cite{Roth16} used deep convolutional neural networks with holistically-nested networks for localization.
Oda et al. \cite{Oda16} used the regression forest technique to localize the pancreas atlas based on global feature values.
Other methods \cite{Roth16,Oda16} showed higher pancreas segmentation accuracies than the previous methods.
Their results provided that an accurate localization technique is crucial for pancreas segmentation.

Pancreas localization also resembles the regression of the pancreas position and size from the information of the surrounding tissues or organs.
Roth et al. \cite{Roth16} performed localization using small 2D patch appearances obtained from around the pancreas.
The 3D information of pancreas surroundings, which has not adequately considered, is important for more accurate localization.
Oda et al. \cite{Oda16} used global feature values for localization.
A global feature value is the first-order difference of the CT values sampled at many local regions in CT volumes.
Global feature values, which only capture the edge information from CT values, do not utilize the more useful information contained in the CT volumes for such localizations as texture or appearance.
In these methods \cite{Roth16,Oda16}, the tissue or organ information surrounding the pancreas is not well utilized for localization.

We propose a pancreas localization and segmentation method from a CT volume utilizing 3D fully convolutional network (FCN) features for localization.
We perform localization using the regression forest technique followed by segmentation with a patient-specific probabilistic atlas.
In the localization, we introduce 3D FCN-based features obtained from a 3D FCN trained to segment abdominal organs including the pancreas.
We use the 3D U-Net \cite{Cicek16} as the 3D FCN structure.
Regression forests perform regression of a pancreas-bounding box from 3D FCN-based features and the first- and second-order intensity difference features calculated in many local regions obtained all over the CT volume.
The 3D FCN features capture statistically useful information for organ localization from a training dataset.
We also introduce a new organ texture feature calculation method from the 3D FCN features.
Since direct use of them at each voxel is ineffective for localization, we introduce a local binary pattern (LBP)-like texture feature calculation method from the 3D FCN features to improve the localization accuracy.
After estimating the pancreas-bounding box, we generate a patient-specific probabilistic atlas and perform pancreas segmentation with it.

The following are the contributions of this paper: (1) introduction of an accurate pancreas localization technique utilizing deep learned features, which are statistically useful for solving organ localization problems, (2) introduction of a new deep learned texture feature measure based on LBP-like feature calculation from deep learned features, and (3) full automation of pancreas segmentation.

\section{Method}

\subsection{Overview}

Our method, which segments the pancreas region from an input CT volume, consists of localization and segmentation processes.
In the localization process, a bounding box of the pancreas is estimated using the regression forest technique.
Regression forests estimate a bounding box from the 3D FCN and the intensity differences features that contain the automatically extracted global features.
After performing the localization process, we generate a patient-specific probabilistic atlas of the pancreas in the bounding box.
Pancreas segmentation is performed using the generated atlas.

\subsection{Pancreas localization}
\label{ssec:localization}

\subsubsection{3D FCN feature extraction:}

We train a 3D U-Net \cite{Cicek16} using a two-stage, coarse-to-fine approach that trains a FCN model to roughly delineate the organs of interest in the first stage: looking at $\sim$40\% of voxels within a simple, automatically generated binary mask of the patient's body.
We then use the predictions of the first-stage FCN to define a candidate region that will be used to train a second FCN.
This step reduces the number of voxels the FCN has to classify to $\sim$10\% while keeping the recall rate high at $>$99\%.
This second-stage FCN can now focus on more detailed organ segmentation.
The last layer of the 3D U-Net contains a $1\times 1\times 1$ convolution that reduces the number of output channels to the number of class labels which is eight in our case: artery, vein, liver, spleen, stomach, gallbladder, pancreas, and background.
We respectively utilize training and validation sets consisting of 281 and 50 clinical CT volumes.
The validation dataset is used to determine good training of the model and avoid overfitting.
This dataset is independent of the testing data in the remainder of this paper.
For feature extraction, we deploy the trained model using a non-overlapping tiles approach on the target volume and extract a 64-dimensional feature vector from the pre-ultimate layer of the 3D U-Nnet for each voxel in the volume.
For computational reasons and to increase the network's field of view when computing each tile, the resolution of all volumes is halved in each dimension.
For implementation, we use the open-source distribution of 3D U-Net\footnote[1]{\url{http://lmb.informatik.uni-freiburg.de/resources/opensource/unet.en.html}} based on the Caffe deep learning library \cite{jia2014caffe}.

\subsubsection{Localization using regression forests:}

\begin{figure}[tb]
  \begin{minipage}{0.48\textwidth}
    \centering
\subfloat[]{\includegraphics[width=0.45\textwidth]{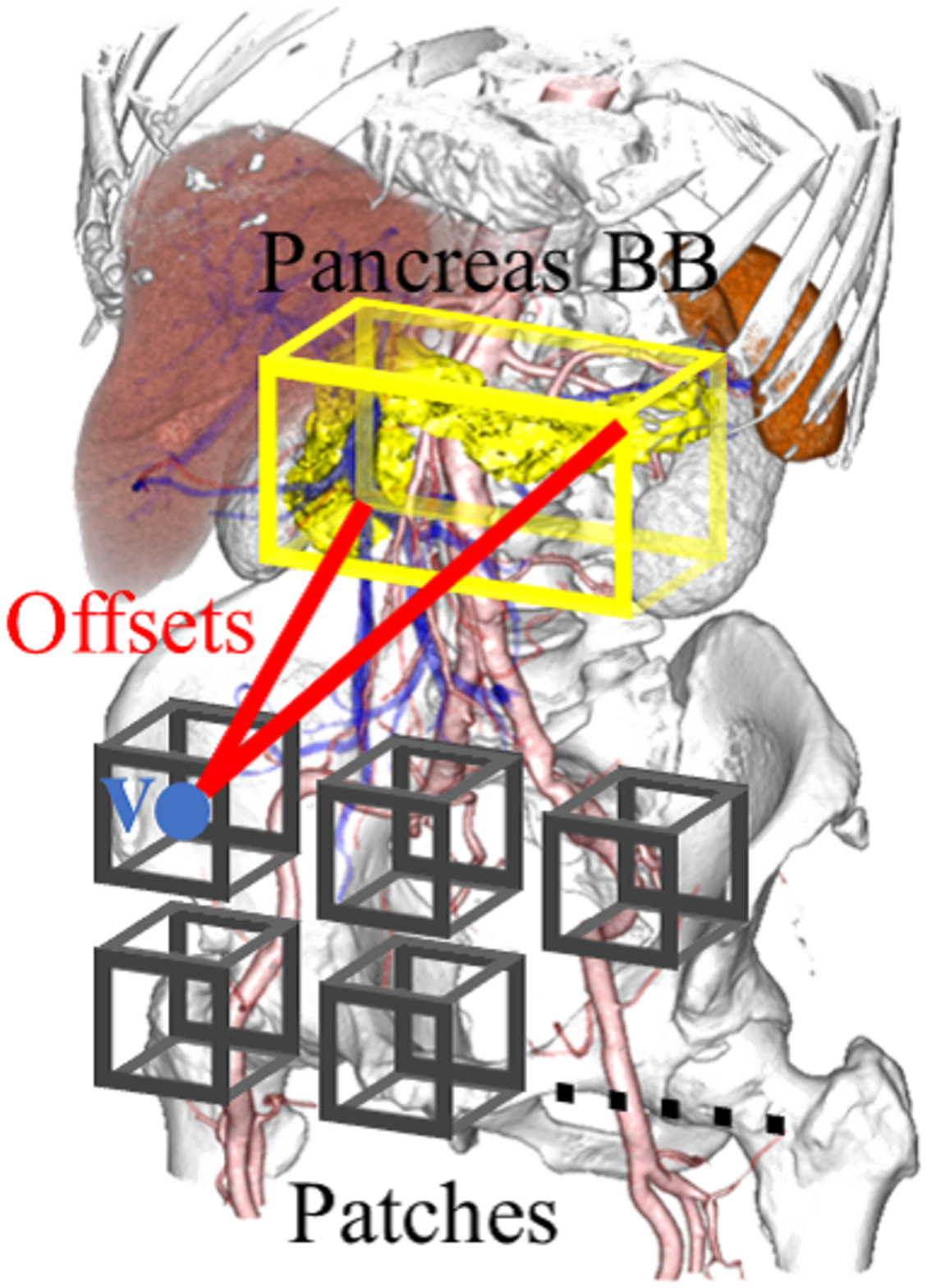}\label{fig:boxdefinition}}
  \end{minipage}
  \begin{minipage}{0.48\textwidth}
    \begin{flushright}
\subfloat[]{\includegraphics[width=0.65\textwidth]{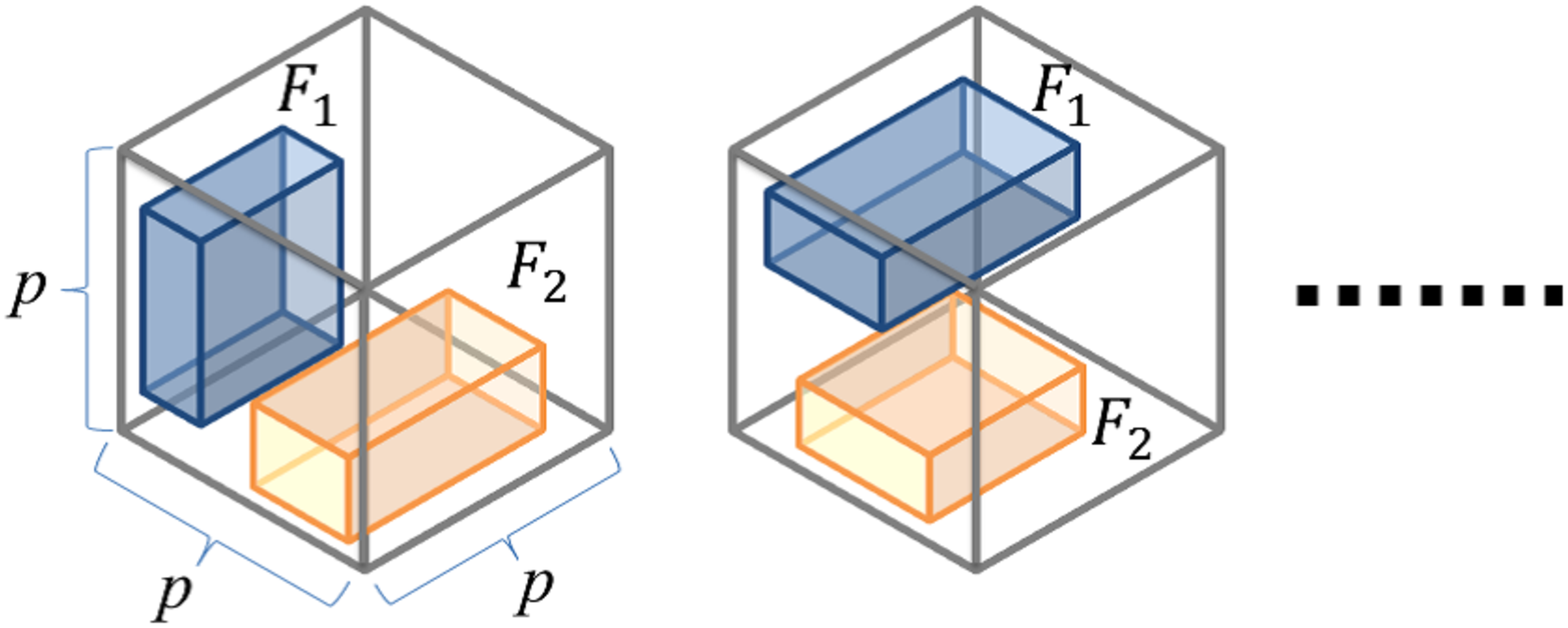}\label{fig:feature_diff1}}

\subfloat[]{\includegraphics[width=0.6\textwidth]{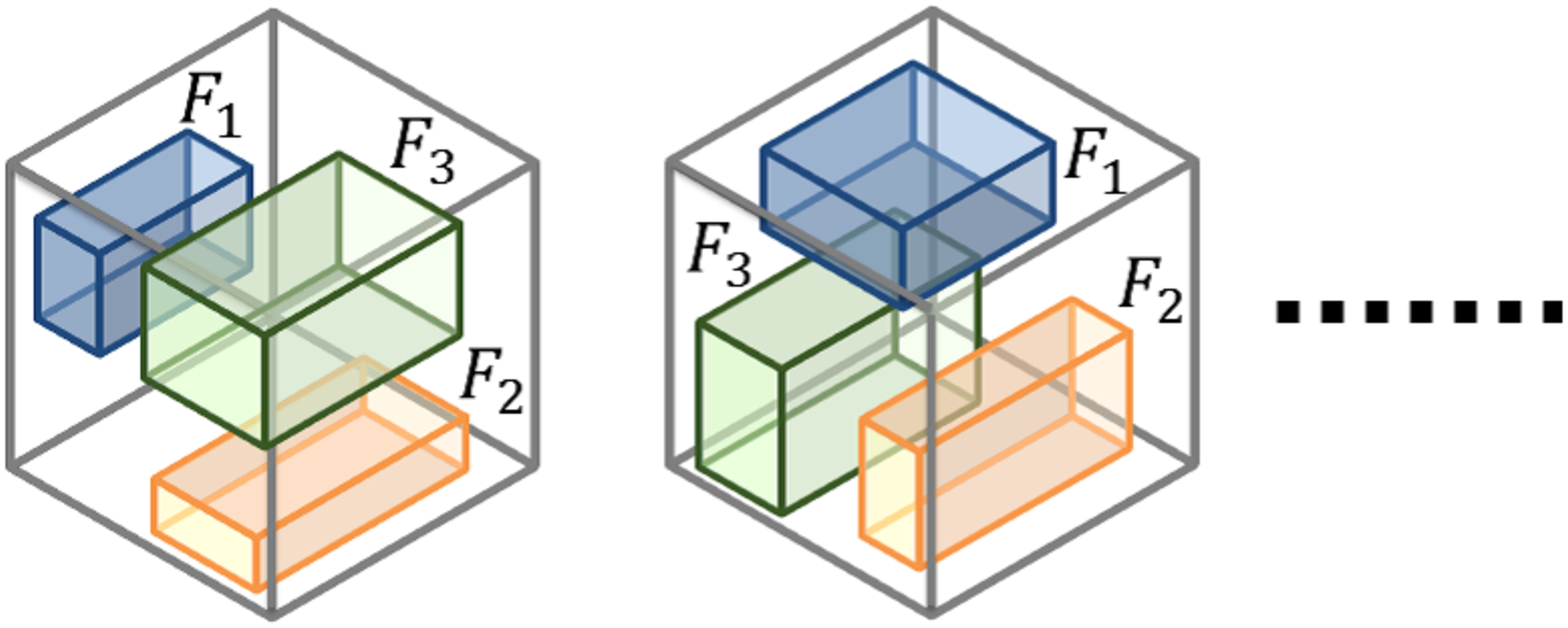}\label{fig:feature_diff2}}
	\end{flushright}
  \end{minipage}\hfill
\caption[]{(a) Schematic illustrations of positions of pancreas-bounding box, patch, and offset of patch in CT volume. (b), (c) Patch and cuboids (represented as cube frames and colored cuboids) used in difference feature value calculation from CT volume: (b) first-order and (c) second-order difference feature calculations.}\label{fig:feature_diff}
\end{figure}

The pancreas-bounding box is axis-aligned with a minimum size that includes a pancreas region, which can be represented as ${\bf b} = (b_{x1}, b_{x2}, b_{y1}, b_{y2}, b_{z1}, b_{z2})$.
We define patches, whose size is $p \times p \times p$ voxels, and the center position is ${\bf v} = (v_{x}, v_{y}, v_{z})$, allocated in the CT volume on a regular grid.
The patch's offset with respect to ${\bf b}$ is represented as ${\bf d} = {\bf b} - {\bf \hat{v}}$, where ${\bf \hat{v}} = (v_{x}, v_{x}, v_{y}, v_{y}, v_{z}, v_{z})$.
They are also shown in Fig. \ref{fig:feature_diff} (a).

\begin{figure}[tb]
\begin{center}
\includegraphics[width=0.53\textwidth]{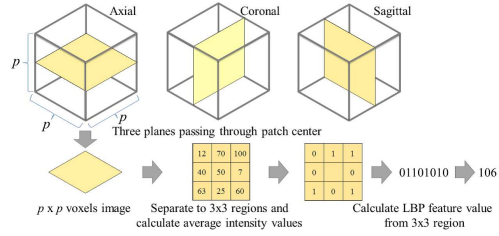}
\caption{Flow of 3D FCN deep learned feature value-based LBP-like feature calculation from patch in feature volume}
\label{fig:feature_unet}
\end{center}
\end{figure}

We use a regression forest to estimate a patch's offset from the feature values calculated in the patches.
A regression forest is constructed to estimate each face of a pancreas-bounding box.
Six regression forests are constructed for a pancreas-bounding box.
The feature values calculated in the patch include: (a) {\it first- and second-order differences of CT values}, (b) {\it 3D FCN deep learned feature value-based LBP-like feature}, and (c) {\it organ likelihoods obtained from 3D FCN}.\\
(a) {\it Differences of CT values}: 
The first- and second-order differences of the CT values in a patch are used as feature values.
The feature value from the first-order difference can be found in a previous work \cite{Oda16}.
We newly introduce the second-order difference, which captures more useful information for localization, such as CT value transitions.
First-order difference feature $f_{1}({\bf v})$ and second-order difference feature $f_{2}({\bf v})$ are represented as
\begin{eqnarray}
f_{1}({\bf v}) &=& \frac{1}{|F_{1}|} \sum_{{\bf q} \in F_{1}} I({\bf q}) - \frac{1}{|F_{2}|} \sum_{{\bf q} \in F_{2}} I({\bf q}),\\
f_{2}({\bf v}) &=& \frac{1}{|F_{1}|} \sum_{{\bf q} \in F_{1}} I({\bf q}) + \frac{1}{|F_{2}|} \sum_{{\bf q} \in F_{2}} I({\bf q}) - \frac{2}{|F_{3}|} \sum_{{\bf q} \in F_{3}} I({\bf q}),
\label{eq:featurevalue}
\end{eqnarray}
where $I({\bf q})$ is a CT value at voxel ${\bf q}$ in a CT volume.
$F_{1}, F_{2},$ and $F_{3}$ are cuboids arbitrarily arranged in a patch.
$|F_{1}|, |F_{2}|,$ and $|F_{3}|$ denote the number of voxels in the cuboids.
The positions and the sizes of the cuboids are randomly selected.
The arrangements of patches and cuboids are shown in Figs. \ref{fig:feature_diff} (b) and (c).\\
(b) {\it 3D FCN deep learned feature value-based LBP-like feature}: 
The deep learned feature values obtained from the last convolution layer of the trained 3D FCN contain useful information for segmentation.
From the 3D U-Net output for each case, we get the feature volume, which contains 64-dimensional feature at each voxel.
The local distribution of the intensity or texture in the feature volume represents crucial information for segmentation.
We calculate the texture representing feature values based on the LBP feature from the feature volume and calculate the LBP-like features of a patch as follows.
We obtain 2D images with the intensity values of the feature volume on three planes, which pass ${\bf v}$ and are parallel to the axial, coronal, and sagittal planes.
One of the three planes is selected randomly.
In the $p \times p$ voxels 2D image, we define the $3 \times 3$ regions of identical size and calculate the average intensity values in each region.
The $3 \times 3$ average intensity values are used to calculate the LBP-like feature of the 2D image of the patch.
The LBP-like feature is used as a feature value of the patch.
This calculation process is shown in Fig. \ref{fig:feature_unet}.\\
(c) {\it Organ likelihoods obtained from 3D FCN}: 
The trained 3D FCN outputs the likelihoods of eight organs (including pancreas) at each voxel.
We use the likelihoods at ${\bf v}$ as a patch's feature values.\\

The regression forest is trained using a training database that includes CT volumes and their corresponding manually segmented pancreas regions.
A regression tree is constructed using patches obtained from the training database, as previously explained \cite{Oda16,Criminisi13}.

In the estimation process of a pancreas-bounding box from an unknown input CT volume, the patches obtained from the CT volume are input to the trained regression trees.
Each regression tree outputs the offset from the patch position.
We get an estimation result of the face of a bounding box as the sum of the offset and the patch position.
The average estimation result of all the patches is the final estimation result of the face of a bounding box.

\subsection{Patient-specific probabilistic atlas generation and pancreas segmentation}

A blood vessel information-based patient-specific probabilistic atlas \cite{Oda16} is quite effective for pancreas segmentation.
We generate this probabilistic atlas using a calculated bounding box.

For the input CT volume, we perform a rough segmentation using the patient-specific probabilistic atlas.
Then precise segmentation is done using the graph-cut method \cite{Boykov01}.
We obtain the final pancreas segmentation results from this precise segmentation process.

\section{Experiments and Discussion}

Our proposed method was evaluated using 146 cases of abdominal CT volumes.
The following are the acquisition parameters of the CT volumes: image size, 512$\times$512 voxels, 263--1061 slices, 0.546--0.820 mm pixel spacing, and 0.40--0.80 mm slice spacing.
The ground truth organ regions were semi-automatically made by three trained researchers.
All of the ground truth regions were checked by a medical doctor.
Our method's parameter was experimentally selected as $p=25$.

We evaluated localization accuracy by measuring the average distance between six faces of estimated and ground truth bounding boxes.
The average distance was 11.4$\pm$4.5 mm when we use only the second-order difference of CT values as regression forest features.
The average distance was improved to 11.0$\pm$4.0 mm when we use all regression forest features proposed in this paper.

\begin{figure}[tb]
  \begin{minipage}{1.0\textwidth}
    \centering
\subfloat[]{\includegraphics[width=0.5\textwidth]{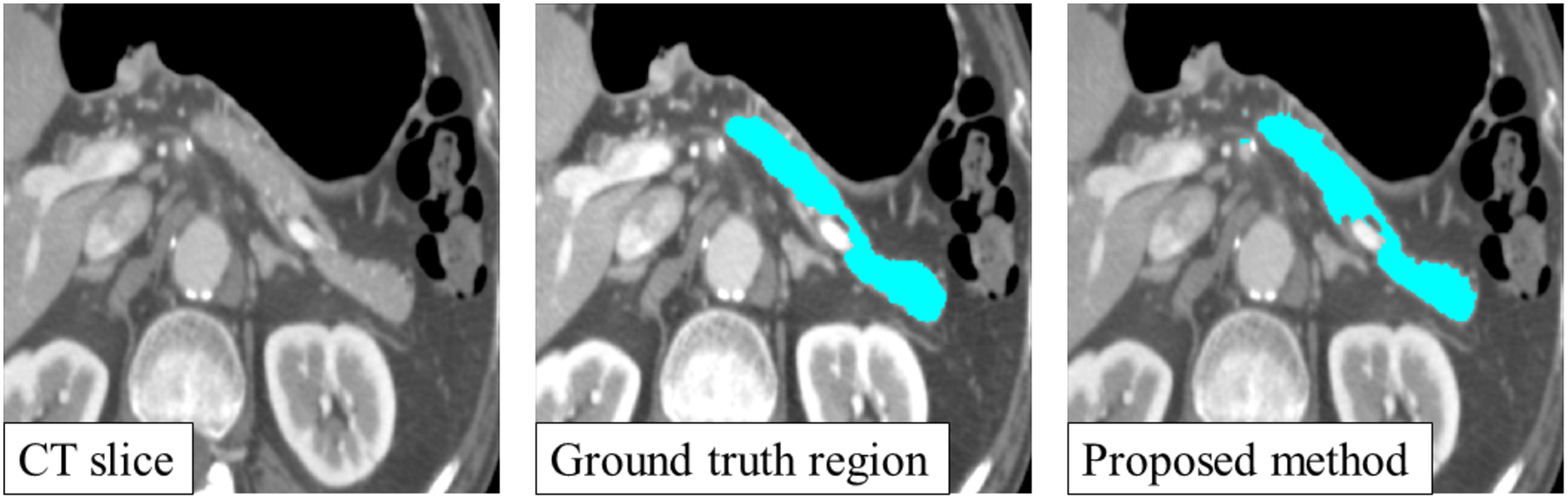}\label{fig:result_a}}
  \end{minipage}
  
  \begin{minipage}{1.0\textwidth}
    \centering
\subfloat[]{\includegraphics[width=0.35\textwidth]{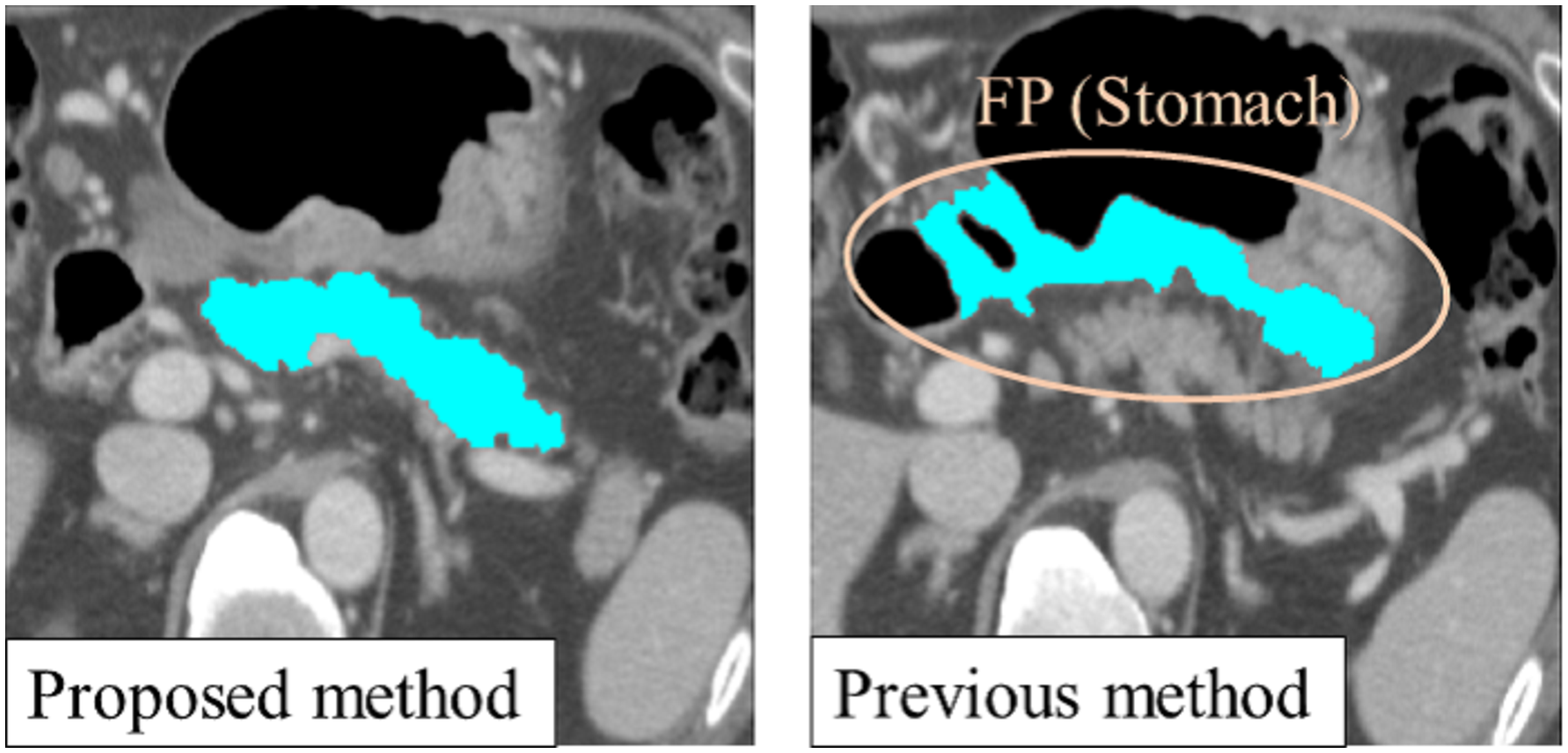}\label{fig:result_b}} \ \ \ \ 
\subfloat[]{\includegraphics[width=0.35\textwidth]{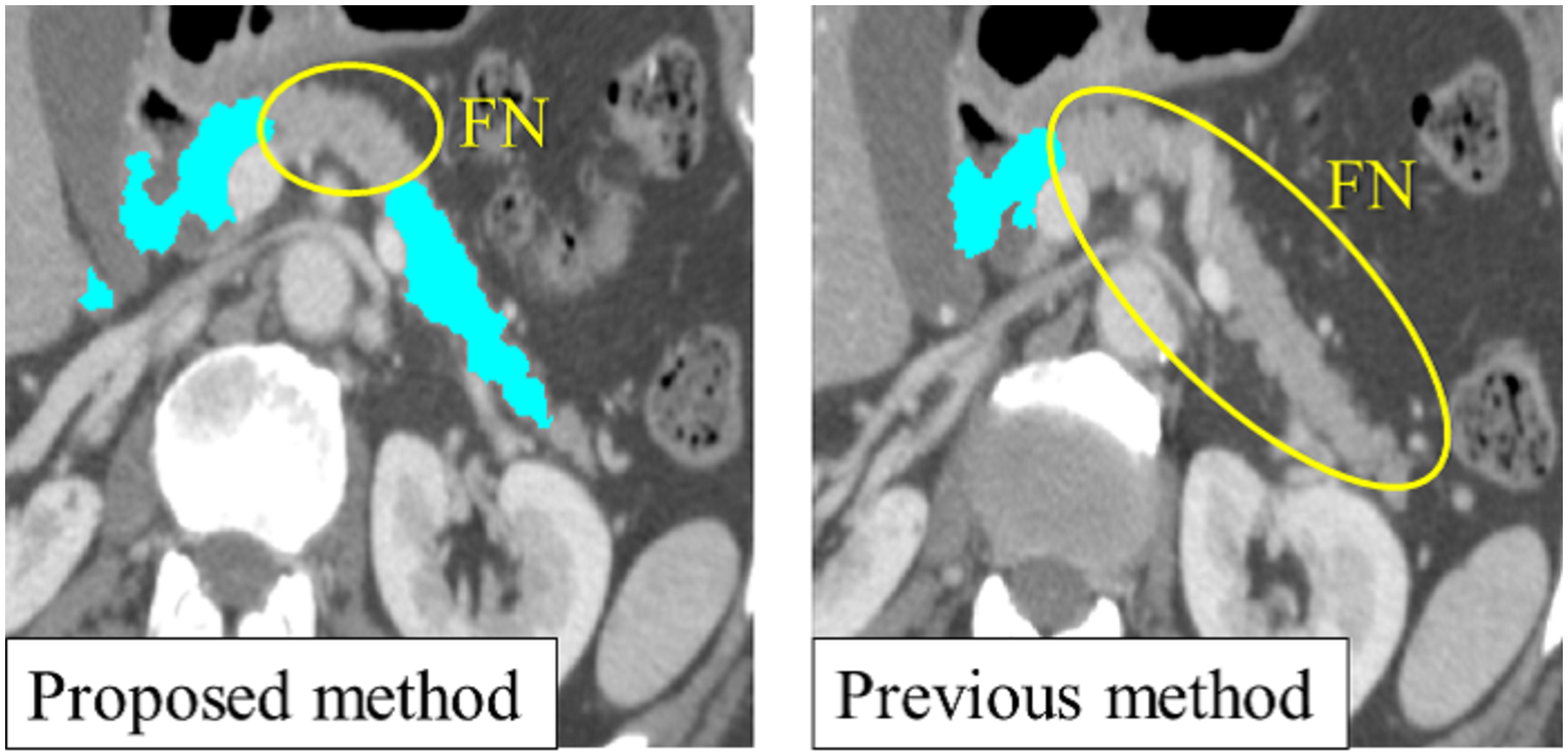}\label{fig:result_c}}
  \end{minipage}\hfill
\caption[]{(a) Pancreas segmentation results on axial slices in estimated bounding boxes. Ground truth and segmented regions are shown as blue regions. (b) and (c) compare segmentation results with previous method \cite{Oda16} using the same cases. DICE improved from 28.0\% to 72.7\% in (b) and 25.4\% to 60.3\% in (c).}\label{fig:result}
\end{figure}

Pancreas segmentation accuracy was evaluated by five-fold cross validation by the following evaluation metrics: Jaccard Index (JI) and Dice Overlap (DICE).
Table \ref{tab:result} compares accuracies of the proposed and previous methods.
Parts of CT volumes used in \cite{Oda16} and our method are same.
Figure \ref{fig:result} shows results.

\begin{table}[tb]
\begin{center}
\caption{Accuracies of proposed and previous pancreas segmentation methods.}
\label{tab:result}
\begin{tabular}{|c|c|c|c|} 
\hline
Method & Data number & JI (\%) & DICE (\%) \\ \hline \hline
Proposed & 146 & 60.6$\pm$16.5 & 73.9$\pm$15.2 \\ \hline
Okada et al.\cite{Okada08} & 86 & 59.2 & 71.8 \\ \hline
Chu et al.\cite{Chu13} & 100 & 54.6$\pm$15.9 & 69.1$\pm$15.3 \\ \hline
Wolz et al.\cite{Wolz13} & 150 & 55.0$\pm$17.1 & 69.6$\pm$16.7 \\ \hline
Karasawa et al.\cite{Karasawa15} & 150 & 61.6$\pm$16.6 & 74.7$\pm$15.1 \\ \hline
Tong et al.\cite{Tong15} & 150 & 56.9$\pm$15.2 & 71.1$\pm$14.7 \\ \hline
Saito et al.\cite{Saito16} & 140 & 62.3$\pm$19.5 & 74.4$\pm$20.2 \\ \hline
Roth et al.\cite{Roth16} & 82 & -- & 78.0$\pm$8.2 \\ \hline
Oda et al.\cite{Oda16} & 147 & 62.1$\pm$16.6 & 75.1$\pm$15.4 \\ \hline
\end{tabular}
\end{center}
\end{table}

We newly introduced the 3D FCN-based features in the regression forest to improve pancreas localization accuracy.
From the experimental result, both of the average and standard deviation of the localization accuracy was improved when we use the new features.
The regression forest estimates a bounding box based on feature values calculated in patches located all over a CT volume.
The difference of CT value feature captures only edge information.
Our new 3D FCN-based features capture more useful information including texture, appearance, and relationships to other organs.
Handcrafted features are useful if they are carefully designed for purpose of use.
Our 3D FCN-based features are automatically designed for organ segmentation.
They also contain useful information for organ localization.
Combinational use of the handcrafted and deep learned features improved the pancreas localization accuracy.

We used the LBP-like feature calculation scheme to extract feature from the 3D FCN feature volume.
Direct use of the voxel value in the feature volume as a feature value for the regression forest is one possible way to perform localization.
Feature value used in the regression forest should contain position-related information because it can be considered as a clue to find a bounding box position.
If we use the feature volume voxel value, many similar values are obtained at many positions in the volume.
Such feature value is not effective for localization.
Our LBP-like feature value captures local texture in the feature volume.
Therefore, calculated feature value is position-related and effective for localization.

We also performed pancreas segmentation using the localization result.
The segmentation accuracy was good among the recent segmentation methods.
Especially, segmentation accuracies were improved among four "difficult to segment" cases whose DICE were lower than 30\% in \cite{Oda16}.
Among them, 28.0\%, 25.4\%, 29.1\%, 18.9\% of DICEs in \cite{Oda16} were improved to 72.7\%, 60.3\%, 35.6\%, 20.1\%, respectively.
Figures \ref{fig:result} (b) and (c) show examples of such cases.
False positives (FPs) and false negatives (FNs) were greatly reduced by the proposed method.
The pancreas touching the surrounding organs in these cases.
The regression forest feature values in \cite{Oda16} were not effective for them.
The proposed deep learned feature values effectively represented localization information even in such cases.
The proposed method accurately localized pancreas and it resulted in reduction of FPs and FNs.

We proposed a pancreas localization method utilizing 3D FCN-based features and intensity difference features.
The regression forest calculates a bounding box of the pancreas from the features.
Also, our method segments pancreas region using a patient-specific probabilistic atlas.
Experimental results using 146 CT volumes showed comparable results to the other state-of-the-art methods.
Future work includes weighting of contribution rates of patches based on their position and application to multi-organ segmentation.

\subsubsection*{Acknowledgments.} Parts of this research were supported by the MEXT/JSPS KAKENHI Grant Numbers 25242047, 26108006, 17H00867, the JSPS Bilateral International Collaboration Grants, and the JST ACT-I (JPMJPR16U9).

\end{document}